\documentclass[runningheads]{llncs}

\usepackage{eccv}

\usepackage{eccvabbrv}

\usepackage{graphicx}
\usepackage{booktabs}
\usepackage{bm}
\usepackage{multirow}
\usepackage{multicol}
\usepackage{hhline}
\usepackage{xcolor}
\usepackage{colortbl} 
\definecolor{Gray}{gray}{0.9}
\usepackage[dvipsnames]{xcolor}
\usepackage{cancel}
\usepackage{comment}
\usepackage{cuted}

\usepackage[accsupp]{axessibility}  

\definecolor{cvprblue}{rgb}{0.21,0.49,0.74}
\definecolor{CornflowerBlue3}{rgb}{0.411, 0.576, 0.749} 

\usepackage{hyperref}

\begin{document}

\title{Learning Probabilistic Embeddings for Unsupervised Action Segmentation} 

\titlerunning{Learning Probabilistic Embeddings for Unsupervised Action Segmentation}

\author{Shuai Li\inst{1} \and
Duc Manh Vu\inst{1} \and
Juergen Gall\inst{1,2}}

\authorrunning{S.~Li et al.}

\institute{University of Bonn, Germany \and
Lamarr Institute for Machine Learning and Artificial Intelligence, Germany
\email{\{lishuai,gall\}@iai.uni-bonn.de}}

\maketitle
\begin{abstract}
  This paper concerns the problem of unsupervised temporal action segmentation for long, untrimmed videos. Recent successful approaches follow a joint representation learning and clustering paradigm, where optimal transport (OT) is adopted to produce pseudo labels for learning frame representations. These approaches alternate between estimating pseudo labels using OT and optimizing the parameters with gradient descent during training, where OT is used for obtaining the final temporal action segmentation. 
  A major limitation of these works is that they learn a deterministic embedding for frame representations. The iterative procedure between learning deterministic embeddings based on pseudo labels and estimating pseudo labels from the learned embedding can thus get quickly stuck in a local optimum. As an alternative, we thus propose to learn a probabilistic embedding for frame representations. The embeddings are modeled by Gaussian distributions and we sample from the distributions before estimating the pseudo labels.  
  We evaluate our approach on several challenging temporal action segmentation datasets and achieve results comparable to, and in some cases, better than the state of the art. Compared to baselines with deterministic embeddings, our approach improves MoF up to 20.7\% and F1-score up to 19.0\%. Our code is available at \url{https://github.com/derkbreeze/PEOT}.
\end{abstract}

\section{Introduction}
\label{sec:intro}

Unsupervised temporal action segmentation is highly relevant for many applications, such as monitoring and optimizing workflows in manufacturing, phenotyping human or animal behavior, as well as human-robot interaction and collaboration~\cite{yang2015robot,ROMEO2025110320}. The task, however, is challenging, as videos can contain different numbers of actions and actions can happen in a different order within the video. 
Furthermore, the same action can occur multiple times within a video. To tackle this problem, approaches based on joint representation learning and clustering~\cite{kumar2022unsupervised, tran2024permutation} have recently become  popular. They iterate during training between estimating pseudo-labels, using optimal transport (OT), and updating the frame embeddings by using the estimated pseudo-labels as target for the cross-entropy loss.
\begin{figure}[!t]
    \centering
    \includegraphics[width=0.9\linewidth]{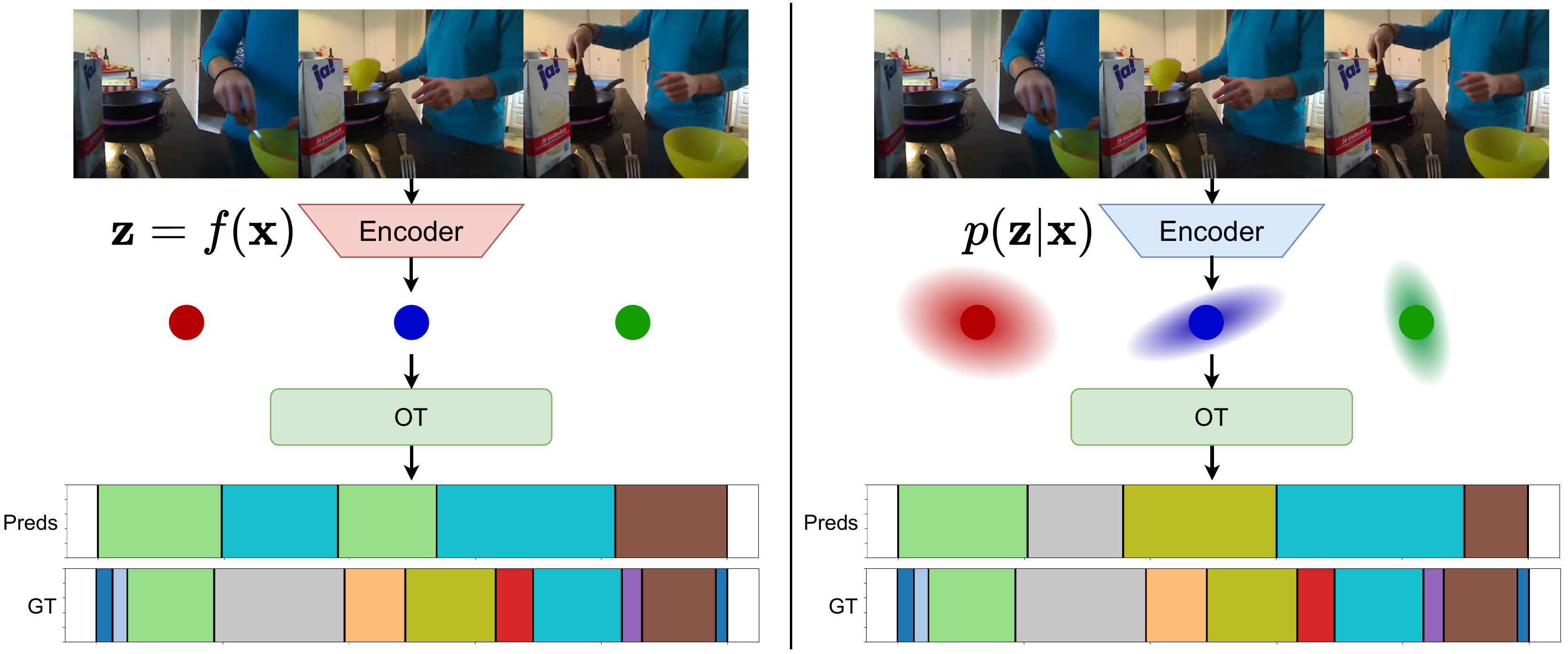}
    \caption{Most previous works learn deterministic embeddings as frame embeddings; we propose to learn probabilistic embeddings such that frame representations are samples from Gaussian distributions that explicitly model embedding uncertainty. Our representations lead to more accurate segmentations compared to the baseline~\cite{xu2024temporally}. Different colors indicate different actions.\label{fig:overview}}
\end{figure}
%
Following this paradigm, Xu and Gould~\cite{xu2024temporally} recently proposed ASOT. The core idea is to use a combination of Kantorovic and Gromov-Wasserstein optimal transport, which offers temporal consistency. They also relax the balanced action assignment assumption in prior works~\cite{kumar2022unsupervised, tran2024permutation} via an unbalanced OT formulation. 
Similar to the idea of~\cite{tran2024permutation}, CLOT~\cite{bueno2025clot} extends ASOT by building a three-level OT that introduces feedback between frame embeddings and action embeddings, which improves the detection of short segments.

We notice that all previous unsupervised works~\cite{kukleva2019unsupervised, vidalmata2021joint, swetha2021unsupervised, li2021action, kumar2022unsupervised, tran2024permutation, bueno2025clot} learn a deterministic embedding as frame representations before computing pseudo-labels. This, however, has the disadvantage that the optimization using optimal transport can get very quickly stuck in a local optimum such that the deterministic embedding overfits to the wrong pseudo-labels. In this work, we thus propose to learn probabilistic frame embeddings using Gaussian distributions as illustrated in~\cref{fig:overview}. We then sample features for each frame from the learned Gaussian distributions, and apply OT to compute the pseudo labels on the sampled features. For estimating the probabilistic frame embeddings, we find that Graph Convolutional Networks (GCN)~\cite{kipf2016semi} perform better than an MLP, as it is commonly used for deterministic frame embeddings, or a TCN~\cite{farha2019ms}.

We evaluate our approach on four challenging unsupervised temporal action segmentation benchmarks, namely   
Breakfast~\cite{kuehne2014language}, Youtube Instructional~\cite{alayrac2016unsupervised}, 50Salads~\cite{stein2013combining}, and Desktop Assembly~\cite{kumar2022unsupervised}, where the results are comparable to, and in some cases, better than the state of the art. More importantly, we demonstrate that the probabilistic frame embeddings improve unsupervised temporal action segmentation compared to a deterministic embedding, using state-of-the-art approaches like ASOT~\cite{xu2024temporally} or VASOT~\cite{ali2025joint} as baselines. Compared to ASOT, our approach improves MoF up to $20.7\%$ ($+12.2$) and F1-score up to $19.0\%$ ($+12.1$). Compared to VASOT, our approach improves MoF up to $16.5\%$ ($+8.8$) and F1-score up to $8.4\%$ ($+5.9$). This shows that probabilistic embeddings are a simple yet efficient approach for improving unsupervised temporal action segmentation.

\vspace{-0.3cm}
\section{Related Work}
\vspace{-0.3cm}
Fully supervised action segmentation approaches yield promising results, but annotating the labels per frame is tedious and time-consuming. While research on weakly-supervised action segmentation requires less labor-intensive effort, unsupervised approaches do not require video labels at all and can scale to large datasets, making them more desirable.

\noindent{\textbf{Unsupervised Action Segmentation.}}
Learning to solve a pretext task is a common paradigm within the unsupervised learning literature. Following this trend, Kukleva~\etal~\cite{kukleva2019unsupervised} train a model to predict relative timestamps for representation learning, followed by a K-means clustering and HMM to produce final segmentation. In a similar vein, VidalMata~\etal~\cite{vidalmata2021joint} extend this method to incorporate visual information while the clustering procedure remains the same. Li and Todorovic~\cite{li2021action} further improve the representation learning and use an HMM with a length model to tackle this problem and achieve decent results. As pointed out by Kumar~\etal~\cite{kumar2022unsupervised}, this paradigm separates representation learning from clustering and as a result, yields sub-optimal segmentation results. They therefore propose a joint representation learning and online clustering method, where a temporal optimal transport is used to estimate pseudo-labels during training. However, the performance of such approaches is limited by the strong fixed action ordering assumption. To address this limitation, Xu and Gould~\cite{xu2024temporally} propose a new method for action segmentation called ASOT, which uses a combination of Kantorovic and Gromov-Wasserstein optimal transport for joint representation learning and clustering, the key innovation is that fixed action ordering assumption is no longer enforced, also ASOT does not assume that different actions are evenly distributed across videos by introducing an unbalanced regularization term. This approach yields promising results. More recently, Elena and Dimiccoli~\cite{bueno2025clot} extend ASOT by introducing the feedback between frame and segment representations. However, the method contains optimal transport on three levels, which makes the model overly complex. In contrast, we adopt a single optimal transport while focusing on learning a better frame representation by modeling uncertainty within representation learning.

\noindent{\textbf{Probabilistic Embeddings.}}
Probabilistic embeddings have been studied in previous works. Vilnis and McCallum~\cite{vilnis2014word} use Gaussian embeddings for word representation learning, and the KL divergence between two Gaussians is used as a similarity measure between embeddings. Oh~\cite{oh2018modeling} describes a similar idea for learning image embeddings, where the model is trained using a variational information bottleneck objective, in order to handle occlusion in images. Shi and Jain~\cite{shi2019probabilistic} propose probabilistic face embeddings that improve face recognition performance. Sun~\etal~\cite{sun2020view} follow~\cite{oh2018modeling} and advocate the use of probabilistic embeddings to address 3D-2D human pose ambiguity. They use the learned embeddings for human pose retrieval and action alignment. However, all these methods work in the supervised learning setup, which requires manual labels to specify whether a pair of observations is matched or not during training, while we address temporal action segmentation and aim to learn probabilistic embeddings in a completely unsupervised manner.

\noindent{\textbf{Graph Convolutional Network for Video Understanding.}}
Graph Convolutional Networks (GCN)~\cite{kipf2016semi} have been widely used for video understanding. Wang~\cite{wang2018videos} describes a spatial-temporal GCN to model object interactions for action recognition, where 2D bounding boxes across frames serve as nodes and edges model their spatial-temporal interaction. Zeng~\cite{zeng2019graph} adopts GCN for action detection. In the context of temporal action segmentation, Huang~\etal~\cite{huang2020improving} propose a classification and a regression GCN, where initial action segments are treated as nodes of the graph, which successfully improves the baseline segmentation performance~\cite{farha2019ms}, particularly in ego-centric cases. More recently, Khan~\etal~\cite{khan2022timestamp} use a GCN to address timestamp-supervised action segmentation. Our work on the other hand, tackles a more challenging unsupervised temporal action segmentation problem by using a GCN to learn smooth frame representations and uncertainty, which has not been done before.

\vspace{-0.4cm}
\section{Approach} \label{sec:approach}
\noindent{\textbf{Problem Formulation.}}
Given a dataset $\mathcal{D} := \{\mathcal{V}_b\}_{b=1}^{B}$ consisting of $B$ videos, frame-wise embeddings $\mathbf{X}\in \mathbb{R}^{N \times D}$ are extracted by an MLP for each video $\mathcal{V}_b$, where $N$ is the number of video frames and $D$ the dimension of the embedding. 
We aim to learn new embeddings $\mathbf{Z} \in \mathbb{R}^{N \times D}$, as well as a set of $K$ action prototypes, represented as $\mathbf{A}:=[\mathbf{a}_{1},\dots, \mathbf{a}_{K}]\in \mathbb{R}^{K \times D}$, where $\mathbf{a}_i\in\mathbb{R}^D$ corresponds to the $i$-th action prototype. In this section, we briefly revisit the optimal transport (OT) formulation~\cite{xu2024temporally} for unsupervised action segmentation, then detail our proposed learning pipeline in~\cref{sec:pe}.

\vspace{-0.4cm}
\subsection{Optimal Transport Formulation} \label{sec:ot}

Our work uses the same OT proposed by Xu and Gould~\cite{xu2024temporally}, which is a combination of Kantorovic Optimal Transport (KOT)~\cite{thorpe2018introduction} and Gromov-Wasserstein Optimal Transport (GWOT)~\cite{peyre2016gromov}.

\noindent{\textbf{Kantorovic Optimal Transport.}} The classical Kantorovich optimal transport is essentially a linear program. Given the cost matrix $\mathbf{C}^k \in \mathbb{R}^{N \times K}_{+}$ and $\mathbf{p} = \frac{1}{N}\mathbf{1}_N $ and $\mathbf{q} = \frac{1}{K}\mathbf{1}_K$, where $\mathbf{1}_N$ and $\mathbf{1}_K$ are $N$ and $K$ dimensional vectors of ones, the optimization aims to find the minimum assignment $\mathbf{T}^\ast$:

\begin{equation} \label{eq:kot_objective}
\mathop{\min}_{\mathbf{T} \in \mathcal{T}} \mathcal{F}_{\text{KOT}}(\mathbf{C}^{k},\mathbf{T}):= \langle \mathbf{C}^k, \mathbf{T} \rangle,
\end{equation}

\begin{equation} \label{eq:kot_constraints}
 \begin{Bmatrix} \mathbf{T}\in \mathbb{R}_{+}^{N\times K}: \mathbf{T}\mathbf{1}_{K} = \mathbf{p}, \mathbf{T}^{\top}\mathbf{1}_{N} = \mathbf{q}
\end{Bmatrix},
\end{equation}
where $\mathcal{T}$ represents a set of possible transportation polytopes. In the context of temporal action segmentation, $\mathbf{C}^k$ can be interpreted as the cost of assigning $N$ frames to $K$ actions.

\noindent{\textbf{Gromov-Wasserstein Optimal Transport.}} The Gromov-Wasserstein optimal transport extends the Kantorovich formulation by allowing for the comparison of histograms defined over incomparable spaces. Given two metric-measure pairs $(\mathbf{C}^v, \mathbf{p})$ and $(\mathbf{C}^a, \mathbf{q})$, the objective function is defined as:
\begin{equation} \label{eq:gwot_objective}
\mathcal{F}_{\text{GW}}(\mathbf{C}^{v},\mathbf{C}^{a},\mathbf{T}):= \sum_{i,k\in[n], j,l\in[m]} L(c^{v}_{ik},c^{a}_{jl})t_{ij} t_{kl},
\end{equation}
where $L:\mathbb{R}\times \mathbb{R}\to \mathbb{R}$ measures the discrepancies between the cost matrices, and $[n]$ and $[m]$ denote the set of video frames and action embeddings.

In order to model the long-tail nature of action segments, ASOT~\cite{xu2024temporally} further relaxes the second hard constraint in~\cref{eq:kot_objective} into a soft constraint by minimizing the KL divergence between action marginals and a uniform action distribution across frames, where a small $\lambda$ encourages a more unbalanced solution $\mathbf{T}$~\cite{chizat2018scaling, de2023unbalanced}. Finally, an entropy regularization term is added to the objective. Therefore, the final OT objective is defined as:
\begin{align} \label{eq:asot_objective}
\nonumber \mathop{\min}_{\mathbf{T} \in \mathcal{T}} \quad & \alpha \mathcal{F}_{\text{GW}}(\mathbf{C}^{v},\mathbf{C}^{a},\mathbf{T}) + (1-\alpha) \mathcal{F}_{\text{KOT}}(\mathbf{C}^k,\mathbf{T}) + \\
 & \lambda D_{\text{KL}}(\mathbf{T}^{\top}\mathbf{1}_N \parallel \mathbf{q}) -\epsilon H(\mathbf{T}),
\end{align}
with the solution satisfying \{$\mathbf{T}\in \mathbb{R}_{+}^{N\times K}: \mathbf{T}\mathbf{1}_K = \mathbf{p}$\}. This non-convex optimization problem is solved using the efficient mirror descent~\cite{xu2024temporally, peyre2016gromov} algorithm with time complexity $\mathcal{O}(NK)$ per iteration.

\noindent{\textbf{Cost Matrices.}}
\cref{eq:asot_objective} contains a set of cost matrices, $\mathbf{C}:=\{ \mathbf{C}^k,\mathbf{C}^{v},\mathbf{C}^{a}\}$ for the KOT and GWOT problem. Specifically, $\mathbf{C}^{k}$ is the visual component and is defined as $c^{k}_{ij}=  1 - \frac{\mathbf{x}_{i}^{\top}\cdot \mathbf{a}_j}{\left\| \mathbf{x}_i\right\| \left\| \mathbf{a}_j\right\|} + \rho \cdot r_{ij} $ and $\mathbf{R} \in \mathbb{R}^{N \times K}$ is the temporal prior commonly used in \cite{kumar2022unsupervised, xu2024temporally} defined as $r_{ij} = |i/N- j/K|$ for $\rho\geq 0$. 

The cost matrices $\mathbf{C}^{v}\in \mathbb{R}^{N\times N}$ and $\mathbf{C}^{a}\in \mathbb{R}^{K\times K}$ serve as structural prior by penalizing associating adjacent frames $(|i - k| \leq Nr, i\neq k)$ to different actions $(j \neq l)$ for two different assignments $t_{ij}$ and $t_{kl}$. However, no penalty is applied to assignments outside the temporal radius $Nr$ or if adjacent frames are mapped to the same action $(j = l)$.

\begin{figure*}[!t]
    \centering   
    \includegraphics[width=0.95\linewidth]{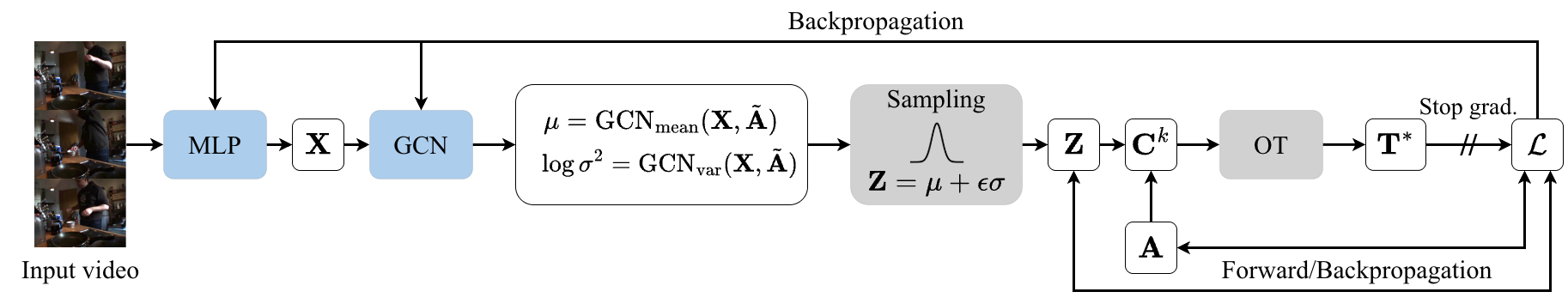}
    \caption{Pipeline of our training scheme. Features of an input video are fed through an MLP to obtain per-frame embeddings $\mathbf{X}$. A temporally weighted graph is constructed, and the normalized adjacency matrix $\mathbf{\tilde{A}}$ along with frame embeddings $\mathbf{X}$ are fed into a Graph Convolutional Network (GCN), which produces the mean and covariance of the frame embeddings. The re-parametrization trick is used during training to obtain probabilistic frame embeddings $\mathbf{Z}$, along with trainable action embeddings $\mathbf{A}$, to produce the cost $\mathbf{C}^k$ for OT that finds the pseudo-label $\mathbf{T}^\ast$. We optimize the cross-entropy loss between the framewise probability distribution and the pseudo-label $\mathbf{T}^\ast$. \textcolor{CornflowerBlue3}{\textbf{Blue}} boxes indicate the \textcolor{CornflowerBlue3}{\textbf{architectural components}} of the model that are trainable, and arrows denote computation/gradient flow.}\label{fig:pipeline}
\end{figure*}

\subsection{Probabilistic Embeddings}\label{sec:pe}
Prior unsupervised segmentation works~\cite{kumar2022unsupervised,tran2024permutation,xu2024temporally,bueno2025clot} essentially learn deterministic embeddings as frame representations, which are used to obtain pseudo-labels via OT. However, networks trained with deterministic embeddings can quickly overfit to the data.
We argue that it is more natural to learn a probabilistic embedding rather than its deterministic counterpart by incorporating the data-dependent (heteroscedastic) uncertainty~\cite{kendall2017uncertainties} into the embeddings. To this end, we propose to parametrize the embeddings such that they follow a Gaussian distribution with a diagonal covariance matrix:
\begin{equation}
    p(\mathbf{z} | \mathbf{x}_i) =  \mathcal{N}(\boldsymbol{\mu}_i, \mathrm{diag}(\boldsymbol{\sigma}_i^2)).
\end{equation} 
where $\boldsymbol{\mu}_i$ and $\boldsymbol{\sigma}_i$ are both $D$-dimensional vectors predicted by the network for the $i$-th frame $\mathbf{x}_i$. In particular, we sample a standard Gaussian noise $\epsilon$ during training and use the differentiable re-parametrization trick~\cite{kingma2013auto} to sample the frame embeddings in the forward pass, \ie, $\mathbf{z}_i = \boldsymbol{\mu}_i + \epsilon \boldsymbol{\sigma}_i$. The sampled embeddings $\mathbf{z}$ are then used to estimate pseudo-labels from OT. Due to the sampling, more variations of the pseudo-labels are generated during training. While the loss is at the beginning higher compared to a deterministic embedding, it converges in general to a better solution as shown in the supplementary material. When the embedding is trained, we use only $\boldsymbol{\mu}_i$ for inference, \ie, the probabilistic embedding does not increase the inference time compared to an approach with a deterministic embedding.           

\vspace{-0.4cm}
\subsubsection{Predicting Probabilistic Embeddings}
In order to integrate temporal context into probabilistic modeling, we propose to use Graph Convolutional Networks (GCN)~\cite{kipf2016semi} to learn smooth representations and uncertainty as shown in \cref{fig:pipeline}. Given an input video $\mathcal{V}$, we first feed it into an MLP $\phi$ to get per-frame embeddings $\mathbf{X} \in \mathbb{R}^{N \times D}$, where $D$ is the dimension of the embedding. Inspired by~\cite{sarfraz2021temporally, wang2018videos}, we construct a temporally weighted graph $\mathcal{G} = (\mathcal{V, E})$ where each frame represents a node and the edges connect every two adjacent frames. The edge weight $a_{ij}$ in the adjacency matrix $\mathbf{A}_{\text{adj}} \in \mathbb{R}^{N \times N}$ is defined as the cosine similarity of pairwise features 
\begin{equation} \label{eq:weightGCN}
a_{ij} = \frac{\phi(\mathbf{x}_i)^T \phi(\mathbf{x}_j)}{||\phi(\mathbf{x}_i)||^2 
||\phi(\mathbf{x}_j)||^2}.
\end{equation}
We show in the ablation study that using a weighted adjacency matrix works better than using an unweighted adjacency matrix. We then add self-connections to the adjacency matrix, \ie, $\hat{\mathbf{A}} = \mathbf{A}_{\text{adj}} + \mathbf{I}$ where $\mathbf{I} \in \mathbb{R}^{N \times N}$ is the identity matrix. This self-connected adjacency matrix is further normalized via $\mathbf{\tilde{A}} = \mathbf{\hat{D}}^{-\frac{1}{2}} \hat{\mathbf{A}} \mathbf{\hat{D}}^{-\frac{1}{2}}$ where $\mathbf{\hat{D}}$ is the degree matrix of $\hat{\mathbf{A}}$. In particular, given the input representation $\mathbf{X}$ and the normalized adjacency matrix $\tilde{\mathbf{A}}$, the output of one GCN layer is computed as:

\begin{equation} \label{eq:gcn}
    \mathbf{Z} = \sigma(\mathbf{\tilde{A}XW})
\end{equation}
where $\mathbf{W} \in \mathbb{R}^{D \times D}$ denotes the weights of the GCN network while $\sigma$ is the activation function and $\mathbf{Z} \in \mathbb{R}^{N \times D}$. GCN makes it possible for the model to incorporate inductive bias in videos. As a result, we use GCN to predict the probabilistic embeddings:

\begin{equation} \label{eq:gcn_uncertainty}
 \mathbf{\boldsymbol{\mu}}= \text{GCN}_{\text{mean}}(\mathbf{X}, \mathbf{\tilde{A}}), \mathbf{\log \boldsymbol{\sigma}}^2 = \text{GCN}_{\text{var}} (\mathbf{X}, \mathbf{\tilde{A}}), 
\end{equation}
where $\boldsymbol{\mu} \in \mathbb{R}^{N \times D}$ and $\boldsymbol{\sigma} \in \mathbb{R}^{N \times D}$ indicate the mean and covariance of the predicted Gaussian distribution for each frame. The network estimates the $\log$ of the variance to ensure positivity. We then use 
$\mathbf{Z} = \boldsymbol{\mu} + \epsilon \boldsymbol{\sigma}$ during training where $\epsilon$ is the noise sampled from standard Gausssian, to estimate pseudo-labels via OT. We will show in the ablation studies that computing $\mathbf{Z}$ using a GCN leads to better embeddings than using an MLP or TCN.

\vspace{-0.3cm}
\subsubsection{Training.} Previous works~\cite{kumar2022unsupervised, tran2024permutation,
xu2024temporally, bueno2025clot} adopt a joint representation learning and clustering paradigm, where the methods alternate between using OT to generate pseudo-labels and optimizing the standard cross-entropy loss using gradient descent, following the Expectation-Maximization (EM) algorithm~\cite{dempster1977maximum}. Given a frame $i$ and its deterministic frame embedding $\mathbf{x}_i$, the probability of assigning frame $i$ to action $j$ is defined as $p_{ij} = \frac{\text{exp}(\mathbf{x}_i^T \mathbf{a}_j / \tau)}{\sum_l \text{exp} (\mathbf{x}_i^T \mathbf{a}_l / \tau)}$ where $\tau$ is the temperature. Together with the pseudo-label $\mathbf{T} \in \mathbb{R}^{N \times K}$ derived from OT, the cross-entropy loss is defined as:

\begin{equation} \label{eq:loss}
\displaystyle \mathcal{L}_{\text{ce}} = -\frac{1}{N} \sum_{i=1}^N \sum_{j=1}^K t_{ij} \log p_{ij}.
\end{equation}

However, such modeling assumes a single ``winner-takes-all'' pseudo-label; models can be over-confident with this pseudo-label and may overfit to the noise inherent in the data. In order to address this problem, we further propose to optimize a novel uncertainty-based cross-entropy loss:

\begin{equation} \label{eq:uncertainty_loss}
\mathcal{L}_{\text{uncer}} = \mathbb{E}_{p_{\phi}(\mathbf{Z}|\mathbf{X})} \mathcal{L}_{\text{ce}}.
\end{equation}

By using probabilistic embeddings $\mathbf{Z}$ rather than its deterministic counterparts $\mathbf{X}$, the algorithm essentially considers different pseudo-labels during training, to avoid the model from overfitting to noisy pseudo-labels $\mathbf{T}$. As it is intractable to integrate the expectation in~\cref{eq:uncertainty_loss}, we propose to approximate the uncertainty loss via Monte Carlo sampling. Concretely, we sample $M$ times from the Gaussian distribution during training to obtain the sampled features and the corresponding cross-entropy losses, followed by averaging them to approximate~\cref{eq:uncertainty_loss}:

\begin{equation} \label{eq:uncertainty_loss_mc}
\displaystyle \mathcal{L}_{\text{uncer}} \approx -\frac{1}{MN} \sum_{m=1}^M \sum_{i=1}^N \sum_{j=1}^K 
t^{(m)}_{ij} \log p^{(m)}_{ij}.
\end{equation}

During inference, we simply take $\mathbf{\boldsymbol{\mu}} = \text{GCN}_{\text{mean}}(\mathbf{X}, \tilde{\mathbf{A}})$ as the final frame embeddings $\mathbf{Z} \in \mathbb{R}^{N \times D}$ to calculate the cost in $\mathbf{C}^k$ and obtain the final segmentation using OT as described in~\Cref{sec:ot}.

\vspace{-0.3cm}
\subsubsection{Computational Complexity.} In comparison to prior works~\cite{xu2024temporally, ali2025joint} which relies on a single pseudo label per EM iteration, our approach requires $M$ pseudo labels during each training iteration and our training complexity is thus $\mathcal{O}(MTNK)$ compared to $\mathcal{O}(TNK)$ where $T$ is the number of steps for gradient descent. Since $M$ is small during training, it adds only a very small additional computational cost as shown in the supplementary material.

\vspace{-0.4cm}
\section{Experiments}\label{sec:results}
\subsection{Experimental Setup}
\noindent{\textbf{Implementation Details.}}
Following~\cite{xu2024temporally}, we design the encoder MLP with a single hidden layer and ReLU activation. We use a one-layer GCN with a graph connecting every 3 neighboring frames. For training, we sample $M=3$ times from the Gaussian distributions. Adam optimizer~\cite{kingma2014adam} is adopted with a learning rate of $10^{-3}$ for the representation learning with a weight decay of $10^{-4}$. We use K-means clustering to initialize the action embeddings, where $K$ equals the ground truth number of actions per activity~\cite{kukleva2019unsupervised, li2021action, kumar2022unsupervised, tran2024permutation, xu2024temporally,bueno2025clot}. For each video, we sample 256 frames from uniformly distributed intervals~\cite{kumar2022unsupervised, xu2024temporally}. 
We provide more hyperparameter setting details in the supplementary material.

\noindent{\textbf{Datasets and Features.}} We conduct experiments on four datasets. For each dataset, we use the same pre-extracted features for training and inference, consistent with prior works~\cite{kukleva2019unsupervised, li2021action, kumar2022unsupervised, xu2024temporally,bueno2025clot}.

\begin{itemize}
    \item \textbf{Breakfast (BF)}~\cite{kuehne2014language} is a large-scale dataset that includes 10 cooking activities. Each video spans a few seconds to several minutes, with multiple actions, where actions can be performed in a different temporal order, for example, \texttt{add teabag} can happen before/after \texttt{pour water} action across videos in the $\texttt{Tea}$ activity. Also, repetitive actions such as \texttt{squeeze orange} and \texttt{pour juice} in the $\texttt{Juice}$ activity can occur multiple times within a video, making it particularly challenging for unsupervised action segmentation. This dataset has a total of 48 actions across 1712 videos. We use the IDT~\cite{wang2013action} features.
    \item \textbf{Youtube Instructional (YTI)}~\cite{alayrac2016unsupervised} has 5 activities and each activity has 30 videos. The average per-video length is roughly 2 minutes, with a large portion of frames being background frames. We use the features provided by~\cite{alayrac2016unsupervised}.
    \item  \textbf{50Salads (FS)}~\cite{stein2013combining} comprises 50 videos of actors making salad. Following prior works~\cite{kukleva2019unsupervised, li2021action, xu2024temporally, bueno2025clot}, we evaluate our model on the Eval granularity, which contains 12 action classes, where actions such as \texttt{cut tomato} and \texttt{cut cheese} are treated as a single \texttt{cut} action. IDT~\cite{wang2013action} features are used as input.
    \item  \textbf{Desktop Assembly (DA)}~\cite{kumar2022unsupervised} has 76 videos of actors performing one assembly activity, where each video is roughly 1.5 minutes long. Each actor conducts 22 actions in a fixed temporal order. We use the features shared by~\cite{kumar2022unsupervised}.
\end{itemize}

\noindent{\textbf{Evaluation Metrics.}} We use the evaluation protocol proposed by~\cite{sener2018unsupervised, kukleva2019unsupervised} for evaluating unsupervised action segmentation. We match the predicted segmentations with the ground truth segmentations via the Hungarian algorithm~\cite{kuhn1955hungarian} across all videos of the same activity, as done by~\cite{sener2018unsupervised, kukleva2019unsupervised, li2021action, kumar2022unsupervised, tran2024permutation, xu2024temporally, bueno2025clot}. 

We use three metrics, namely mean-over-frames (MoF), F1-score, and mean intersection-over-union (mIoU). MoF calculates the percentage of correct per-frame predictions. F1-score is defined on the per-segment level, \ie, for a predicted segment that is matched to a ground truth segment. If the number of correct frames exceeds 50\% of the ground truth segment length, it is regarded as a true positive segment~\cite{kukleva2019unsupervised}. mIoU averages IoU over all classes. 
We report F1 and mIoU averaged across all activities for a dataset.

\vspace{-0.4cm}
\subsection{Comparison with State of the Art}

We name our unsupervised segmentation approach as probabilistic embeddings optimal transport (PEOT) and evaluate it on Breakfast~\cite{kuehne2014language}, Youtube Instr.~\cite{alayrac2016unsupervised}, 50Salads (Eval)~\cite{stein2013combining}, as well as Desktop Assembly~\cite{kumar2022unsupervised}. The results are summarized in~\Cref{tab:sota_activity}. 
Our method outperforms the state of the art for 9 out of 12 metrics and datasets. On Breakfast and Desktop Assembly, our approach outperforms the state of the art for all metrics. On Youtube Instr.\ and 50Salads (Eval), the very recent works VASOT~\cite{ali2025joint} and CLOT~\cite{bueno2025clot}, which are complementary extensions of ASOT, report better results for some metrics, whereas our approach achieves the highest MoF. VASOT~\cite{ali2025joint} is a recently proposed method that jointly solves action alignment and action segmentation by extending the fused Gromov-Wasserstein optimal transport to match a pair of videos from the same activity. It surpasses ASOT at the cost of increased computational complexity. Our approach consistently yields better results than VASOT in terms of MoF and F1, despite having a much simpler methodology design. CLOT~\cite{bueno2025clot} is another recent extension to ASOT. It introduces feedback between frame-wise and segment-wise representations via the cross-attention~\cite{vaswani2017attention} mechanism. The system also integrates a projection-based sliced Wasserstein distance~\cite{nguyen2023energy} that successfully detects small segments, a significant limitation of ASOT. On the Breakfast and Desktop Assembly dataset, our approach outperforms CLOT on all metrics and it performs comparable than CLOT on the Youtube Instr.\ and 50Salads (Eval) datasets with lower mIoU and higher MoF.

The gain compared to ASOT~\cite{xu2024temporally} is most important since it is our baseline. MoF compared to ASOT is increased by {+4.6 (8.2\%)} on Breakfast, {+1.5 (4.7\%)} on YTI, {+5.6 (9.4\%)} on 50Salads, and {+0.8 (1.1\%)} on Desktop Assembly, and the F1-score is increased by {+2.2 (5.7\%)} on Breakfast, {+5.3 (16.5\%)} on YTI, {+5.3 (9.9\%)} on 50Salads, and {+7.8 (11.5\%)} on Desktop Assembly.

While we compare in~\Cref{tab:sota_activity} our approach to results that have been reported in the literature, we also provide a direct comparison to ASOT~\cite{xu2024temporally} and VASOT~\cite{ali2025joint} using the public available source code in~\Cref{tab:ablation_vasot}. 

Since the source code of CLOT~\cite{bueno2025clot} is only partially available, we could not include it in the comparison. To demonstrate that probabilistic embeddings improve deterministic embeddings, we added them to ASOT~\cite{xu2024temporally} and VASOT~\cite{ali2025joint} as baselines. In both cases, we observe improvements with the largest gains on the Desktop Assembly dataset. Compared to ASOT, our approach improves MoF up to $20.7\%$ ($+12.2$) and F1-score up to $19.0\%$ ($+12.1$). Compared to VASOT, our approach improves MoF up to $16.5\%$ ($+8.8$) and F1-score up to $8.4\%$ ($+5.9$). This shows that probabilistic embeddings are a simple yet efficient approach for improving unsupervised temporal action segmentation. 
We provide qualitative results in~\Cref{sec:qual}.

 \begin{table*}[!t]
    \caption{Comparisons of action segmentation performance obtained by applying the \textcolor{blue}{Hungarian matching} at the \textcolor{blue}{activity-level} on the Breakfast~\cite{kuehne2014language}, Youtube Instr.~\cite{alayrac2016unsupervised}, 50Salads (Eval)~\cite{stein2013combining} and Desktop Assembly~\cite{kumar2022unsupervised} datasets. The highest accuracy is indicated in \textbf{bold}, and the second highest is \underline{underlined}. }
    \label{tab:sota_activity}
    \centering
    \scriptsize
    \tabcolsep=0.1cm
    \begin{tabular}{ccccccccccccc}
    \toprule
     \multirow{2}{*}{Methods} & \multicolumn{3}{c}{Breakfast} &\multicolumn{3}{c}{YTI} & \multicolumn{3}{c}{50Salads (Eval)} & \multicolumn{3}{c}{DA} \\
    \cmidrule(lr){2-4} \cmidrule(lr){5-7} \cmidrule(lr){8-10} \cmidrule(lr){11-13}
    & MoF & F1 & mIoU & MoF & F1 & mIoU & MoF & F1 & mIoU & MoF & F1 & mIoU  \\
    \midrule
     CTE~\cite{kukleva2019unsupervised} & 41.8 & 26.4 & - & 39.0 & 28.3 & - & 35.5 & - & - & 47.6 & 44.9 & - \\
    VTE~\cite{vidalmata2021joint} & 48.1 & - & - & - & 29.9 & - & 30.6 & - & - & - & - & - \\
    UDE~\cite{swetha2021unsupervised} & 47.4 & 31.9 & - & 43.8 & 29.6 & - & 42.2 & 34.4  & - & - & - & - \\
    ASAL~\cite{li2021action} & 52.5 & 37.9 & - & 44.9 & 32.1 & -  & 39.2 & - & - & - & - & - \\
    TOT~\cite{kumar2022unsupervised} & 47.5 & 31.0 & - & 40.6 & 30.0 & - & 47.4 & 42.8 & - & 56.3 & 51.7 & - \\
    TOT+~\cite{kumar2022unsupervised} & 39.0 & 30.3 & - & 45.3 & 32.9 & - & 44.5 & 48.2 & - & 58.1 & 53.4 & - \\
    UFSA~\cite{tran2024permutation} & 52.1 & 38.0 & - & 49.6 & 32.4 & - & 55.8 & 50.3 & - & 65.4 & 63.0 & - \\
    ASOT~\cite{xu2024temporally} & 56.1 & 38.3 & 18.6 & 52.9 & 32.1 & \underline{24.7} & 59.3 & 53.6 & 30.1 & 70.4 & 68.0 & 45.9 \\
    HVQ~\cite{spurio2025hierarchical} & 54.4 & 39.7 & - & 50.3 & 35.1 & - & - & - & - & - & - & - \\
    VASOT~\cite{ali2025joint} & 57.5 & 39.0 & \underline{18.8} & 53.2 & 35.7 & \textbf{25.2} & \underline{60.6} & 57.4 & \underline{34.5} & \underline{70.9} & \underline{75.1} & \underline{49.3} \\
    CLOT~\cite{bueno2025clot} & \underline{60.1} & \underline{40.1} & 18.5 & \underline{54.4} & \underline{36.7} & 23.4 & 59.4 & \textbf{63.2} & \textbf{38.8} & 68.8 & 72.6 & 48.1 \\
    
    \rowcolor{Gray} PEOT (Ours) & \textbf{60.7} & \textbf{40.5} & \textbf{19.0} & \textbf{55.4} & \textbf{37.4} & 22.9 & \textbf{64.9} & \underline{58.9} & 30.2 & \textbf{71.2} & \textbf{75.8} & \textbf{51.7}\\
    \bottomrule
    \end{tabular}
\end{table*}

\begin{table*}[!t]
    \caption{Effect of probabilistic embeddings on the four datasets for ASOT and VASOT: Breakfast~\cite{kuehne2014language}, Youtube Instr.~\cite{alayrac2016unsupervised}, 50Salads (Eval)~\cite{stein2013combining}, and Desktop Assembly~\cite{kumar2022unsupervised}. The results of the baselines have been computed using the public available source codes.}
    \label{tab:ablation_vasot}
    \centering
    \scriptsize
    \tabcolsep=0.08cm
    \begin{tabular}{l|ccccccccccccc@{}}
    \toprule
     \multicolumn{1}{c}{} & \multicolumn{3}{c}{Breakfast} & \multicolumn{3}{c}{YTI} & \multicolumn{3}{c}{50Salads (Eval)} &
     \multicolumn{3}{c}{DA} \\
     \cmidrule(lr){2-4} \cmidrule(lr){5-7} \cmidrule(lr){8-10} \cmidrule(lr){11-13}
     Methods & MoF & F1 & mIoU & MoF & F1 & mIoU & MoF & F1 & mIoU & MoF & F1 & mIoU  \\
    \midrule
    ASOT~\cite{xu2024temporally} & 56.4 & 35.7 & 17.2 & 49.0 & 34.1 & 23.7 & 59.4 & 56.7 & 25.6 & 59.0 & 63.7 & 40.6 \\
    \rowcolor{Gray}
    ASOT~\cite{xu2024temporally} + Prob. & 60.7 & 40.5 & 19.0 & 55.4 & 37.4 & 22.9 & 64.9 & 58.9 & 30.2 & 71.2 & 75.8 & 51.7 \\
    VASOT~\cite{ali2025joint}  
    & 54.5 & 35.3 & 15.5  & 47.5 & 30.4 & 18.0  & 53.4 & 51.9 & 26.7 & 67.9 & 70.2 & 48.0 \\
    \rowcolor{Gray}
    VASOT~\cite{ali2025joint} + Prob.
    & 57.2 & 36.1 & 15.8 & 52.5 & 32.5 & 18.8 & 62.2 & 53.6 & 26.4 & 71.1 & 76.1 & 51.8 \\
    \bottomrule
    \end{tabular}
\end{table*}

\subsection{Ablation Study}     
\begin{table*}[!t]
    \caption{Impact of probabilistic embedding and GCN on the four datasets: Breakfast~\cite{kuehne2014language}, Youtube Instr.~\cite{alayrac2016unsupervised}, 50Salads (Eval)~\cite{stein2013combining} and Desktop Assembly~\cite{kumar2022unsupervised}.}
    \label{tab:ablation}
    \centering
    \scriptsize
    \tabcolsep=0.11cm
    \begin{tabular}{c|c|ccccccccccccc@{}}
    \toprule
     \multicolumn{2}{c}{} & \multicolumn{3}{c}{Breakfast} &\multicolumn{3}{c}{YTI} & \multicolumn{3}{c}{50Salads (Eval)} & \multicolumn{3}{c}{DA} \\
     \cmidrule(lr){3-5} \cmidrule(lr){6-8} \cmidrule(lr){9-11} \cmidrule(lr){12-14}
     Prob.& GCN & MoF & F1 & mIoU & MoF & F1 & mIoU & MoF & F1 & mIoU & MoF & F1 & mIoU  \\
    \midrule
    & & 56.4 & 35.7 & 17.2 & 49.0 & 34.1 & 23.7 & 59.4 & 56.7 & 25.6 & 59.0 & 63.7 & 40.6 \\
    \checkmark &  & 59.2 & 34.9 & 15.1 & 49.4 & 33.7 & 20.0 & 60.1 & 56.3 & 23.7 & 63.8 & 63.9 & 42.3 \\
    & \checkmark  & 58.2 & 40.7 & 18.5 & 48.9 & 33.9 & 22.9 & 50.4 & 48.7 & 17.2 & 63.8 & 70.5 & 46.2 \\
    \rowcolor{Gray} \checkmark & \checkmark  & 60.7 & 40.5 & 19.0 & 55.4 & 37.4 & 22.9 & 64.9 & 58.9 & 30.2 & 71.2 & 75.8 & 51.7 \\
    \bottomrule
    \end{tabular}
\end{table*}

\begin{table*}[t]
\centering

\begin{minipage}[!t]{0.48\textwidth}
\centering
\scriptsize
\tabcolsep=0.3cm
\caption{Comparison of MLP, TCN, and GCN on Breakfast~\cite{kuehne2014language}.}
\label{tab:ablation_gcn_tcn}
\begin{tabular}{lccc}
\toprule
Methods & MoF & F1 & mIoU \\
\midrule
MLP & 59.2 & 34.9 & 15.1\\
TCN & 59.2 & 38.9 & 17.9\\
\rowcolor{Gray} GCN & 60.7 & 40.5 & 19.0 \\
\bottomrule
\end{tabular}
\end{minipage}
\hfill
\begin{minipage}[!t]{0.48\textwidth}
\centering
\scriptsize
\tabcolsep=0.3cm
\caption{Ablation study of graph connectivity for GCN on Breakfast~\cite{kuehne2014language}.}
\label{tab:ablation_gcn_connectivity}
\begin{tabular}{lccc}
\toprule
Connectivity & MoF & F1 & mIoU \\
\midrule
 5 frames  & 58.9 & 40.0 & 18.5 \\
 \rowcolor{Gray} 3 frames  & 60.7 & 40.5 & 19.0 \\
\bottomrule
\end{tabular}
\end{minipage}
\end{table*}

\begin{table}[!t]
\caption{Impact of weighted adjacency matrix for GCN on Breakfast~\cite{kuehne2014language}.}
\label{tab:ablation_gcn_adj}
\centering
\scriptsize
\tabcolsep=0.7cm
\begin{tabular}{c c c c} 
 \hline
 Methods & MoF & F1 & mIoU \\
 \hline
 Unweighted adjacency matrix & 59.5 & 40.2 & 18.0 \\ 
 \rowcolor{Gray} Weighted (learned) adjacency matrix & 60.7 & 40.5 & 19.0 \\
 \hline
\end{tabular}
\end{table}

\begin{table}[!t]
\caption{Impact of number of samples on Breakfast~\cite{kuehne2014language}.}
\label{tab:ablation_m}
\centering
\scriptsize
\tabcolsep=1.1cm
\begin{tabular}{c c c c} 
 \hline
 Samples & MoF & F1 & mIoU \\
 \hline
 $M=1$ & 60.0 & 39.4 & 18.4 \\
 $M=2$ & 60.3 & 40.1 & 18.8 \\
 \rowcolor{Gray} $M=3$ & 60.7 & 40.5 & 19.0 \\
 $M=5$ & 60.6 & 40.3 & 18.9 \\
 \hline
\end{tabular}
\end{table}

\subsubsection{Impact of Probabilistic Embeddings and GCN.}
We study the impact of different components on our system in~\Cref{tab:ablation}. If we do not use probabilistic embeddings and a GCN to estimate them, our approach is the same as ASOT~\cite{xu2024temporally} since we use the same MLP architecture and OT formulation as ASOT. 
     
If we learn the probabilistic embeddings using an MLP (row 2 of \Cref{tab:ablation}), \ie, by parameterizing $\boldsymbol{\mu}$ and  $\log \boldsymbol{\sigma}^2$ of the Gaussian distributions with one added MLP layer, it consistently improves MoF on all datasets. However, it does not improve F1 and mIoU, it even decreases these metrics in most cases. Since the MLP estimates the Gaussian distributions using the features of only one frame, the estimates are not very reliable. If we learn the probabilistic embeddings using a GCN (row 4) instead of an MLP, we get except of mIoU on YTI a substantial improvement for all metrics and datasets compared to the baseline (row 1). The MoF is improved by $+4.3$ (7.6\%), $+6.4$ (13.1\%), $+5.5$ (9.3\%), and $+12.2$ (20.7\%) for the Breakfast, YTI, 50Salads, and Desktop Assembly datasets, respectively. The F1-score is improved by $+4.8$ (13.4\%), $+3.3$ (9.7\%), $+2.2$ (3.9\%), and $+12.1$ (19.0\%). In particular the improvements over the baseline for MoF and F1-score are very large. We show that these improvements are not only due to the GCN (row 3). While the GCN improves the baseline as well in many cases, the largest and most consistent improvements are due to the probabilistic embeddings. 

In~\Cref{tab:ablation_gcn_tcn}, we also evaluate the impact of replacing the GCN layer by a TCN~\cite{farha2019ms} layer, \ie, a 1-dimensional temporal convolution of kernel size 3. While TCN performs better than an MLP, the GCN layer performs best. While TCN uses a convolution kernel with fixed weights over all temporal frames, the GCN layer adaptively adjusts the weights based on the input features.

\begin{table}[!t]
    \caption{Comparison with other stochastic regularization approaches: Gauss: GCN+fix Gaussian noise, Drop: GCN+Dropout, Ours: GCN+Learned uncertainty.}
    \label{tab:ablation_stoch}
    \centering
    \scriptsize
    \tabcolsep=0.14cm
    \begin{tabular}{l|ccccccccccccc@{}}
    \toprule
      & \multicolumn{3}{c}{Breakfast} &\multicolumn{3}{c}{YTI} & \multicolumn{3}{c}{50Salads (Eval)} & \multicolumn{3}{c}{DA} \\
     \cmidrule(lr){2-4} \cmidrule(lr){5-7} \cmidrule(lr){8-10} \cmidrule(lr){11-13}
     Stoch. & MoF & F1 & mIoU & MoF & F1 & mIoU & MoF & F1 & mIoU & MoF & F1 & mIoU  \\
    \midrule
    Gauss & 58.3 & 38.7 & 18.1 & 53.5 & 34.9 & 21.3 & 61.0 & 58.8 & 30.2 & 61.9 & 63.2 & 42.0 \\
    Drop & 58.5 & 40.7 & 18.4 & 48.5 & 34.3 & 22.3 & 62.0 & 58.6 & 29.6 & 65.3 & 73.8 & 48.6 \\
     \rowcolor{Gray} Ours & 60.7 & 40.5 & 19.0 & 55.4 & 37.4 & 22.9 & 64.9 & 58.9 & 30.2 & 71.2 & 75.8 & 51.7 \\
    \bottomrule
    \end{tabular}%
    \vspace{-5mm}
\end{table}

\vspace{-0.5cm}
\subsubsection{Impact of Graph Connectivity in GCN.}
We investigate the impact of adding more connections to build the graph $\mathcal{V}$ for the GCN. By default, each node is connected to three nodes, \ie, the node at frame $i$ is connected to the nodes at frames $i-1$, $i$, and $i+1$. We tried more connections such that the node at frame $i$ is connected to the nodes at frames $i-2$ and $i+2$ as well, but the results in~\Cref{tab:ablation_gcn_connectivity} show that increasing the number of connections does not increase the performance. This might be attributed to the fact that adding more edges in the graph leads to the over-smoothing problem in GCNs~\cite{kipf2016semi, zeng2019graph}. For this reason, we use the 3-frames connection graph for the GCN.

\vspace{-0.5cm}
\subsubsection{Impact of Weighted Adjacency Matrix in GCN.}
We investigate the benefit of using a weighted adjacency matrix \eqref{eq:weightGCN} in our GCN in~\Cref{tab:ablation_gcn_adj}. For comparison, we use a GCN layer without weighted adjacency matrix, \ie, $a_{ij}$ in $\mathbf{A}_{\text{adj}}$ is 1 if there is a connection between frame $i$ and $j$ and 0 otherwise. We notice that using a weighted adjacency matrix for the GCN leads to better performance.

\vspace{-0.5cm}
\subsubsection{Impact of $M$.}
We study the effect of using a different number of Monte Carlo samples during training~\eqref{eq:uncertainty_loss_mc} in~\Cref{tab:ablation_m}. Note that for $M=1$ we still have a probabilistic embedding, but we sample only once from the learned Gaussian distributions.    
We notice that increasing $M$ from 1 to 3 increases the performance, but it saturates after $M=3$.

\vspace{-0.5cm}
\subsubsection{Comparisons of Probabilistic Embeddings with Stochastic Regularizations.}
We compare to stochastic regularization approaches in~\Cref{tab:ablation_stoch}, \ie, adding Gaussian noise to deterministic embeddings or adding dropout to the GCN. All methods use the same GCN architecture for fair comparison. Except of F1 on Breakfast, where dropout performs slightly better, our approach outperforms the stochastic regularization methods. For instance, F1 is $+2.5$ (7.2\%) higher on YTI and MoF is $+5.9$ higher (9\%) on DA. 
It demonstrates that our approach is more effective than other stochastic regularization approaches.

\begin{figure*}[!t]
\centering
\begin{subfigure}[t]{0.48\textwidth}
    \centering
    \includegraphics[width=\linewidth]{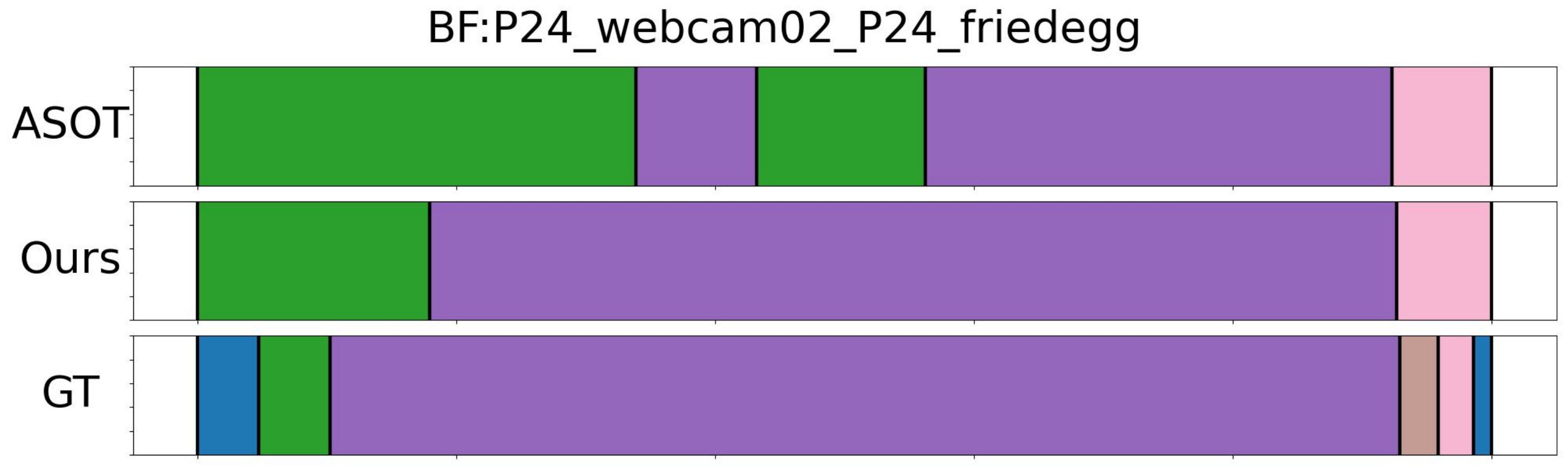}
    \caption{Breakfast~\cite{kuehne2014language}}
\end{subfigure}
\hfill
\begin{subfigure}[t]{0.48\textwidth}
    \centering
    \includegraphics[width=\linewidth]{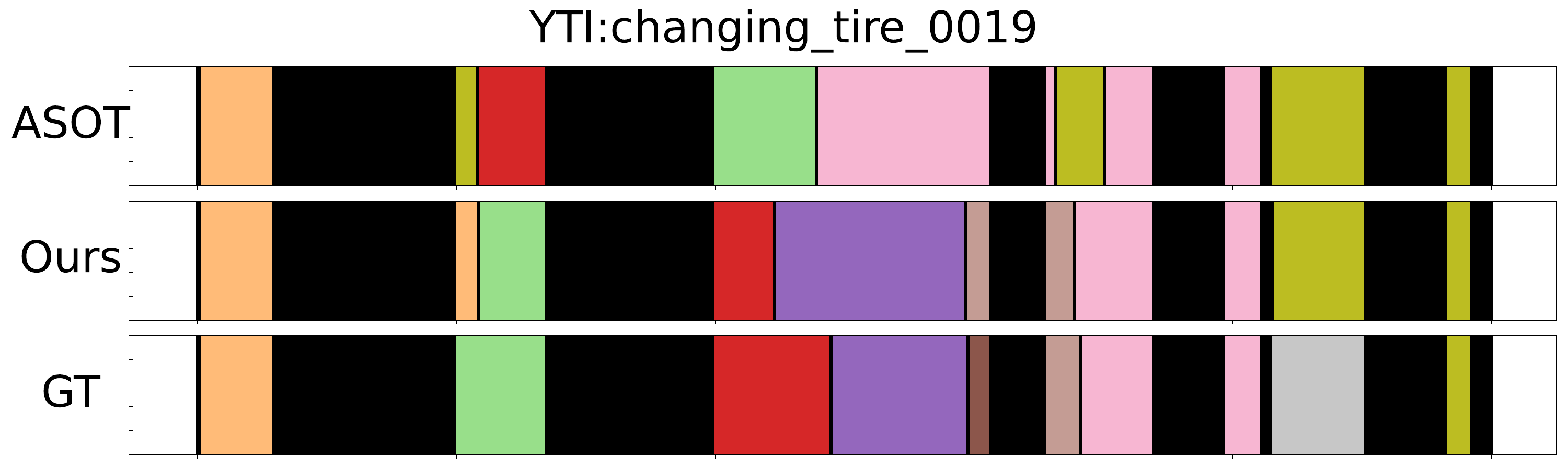}
    \caption{Youtube Instructional~\cite{alayrac2016unsupervised}}
\end{subfigure}

\begin{subfigure}[t]{0.48\textwidth}
    \centering
    \includegraphics[width=\linewidth]{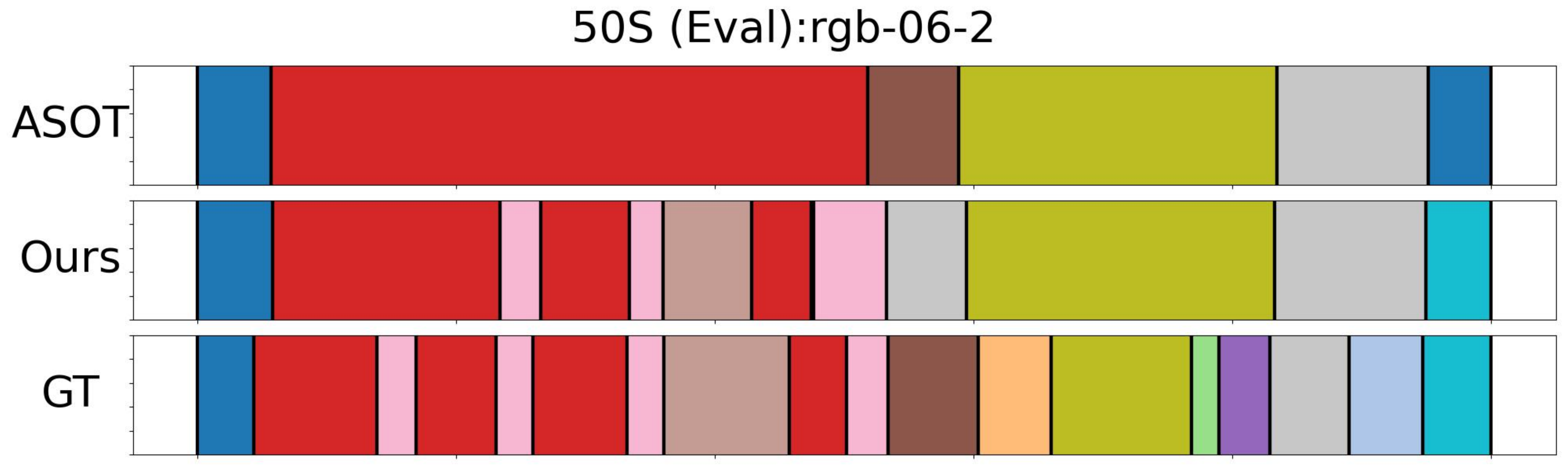}
    \caption{50Salads~\cite{stein2013combining}}
\end{subfigure}
\hfill
\begin{subfigure}[t]{0.48\textwidth}
    \centering
    \includegraphics[width=\linewidth]{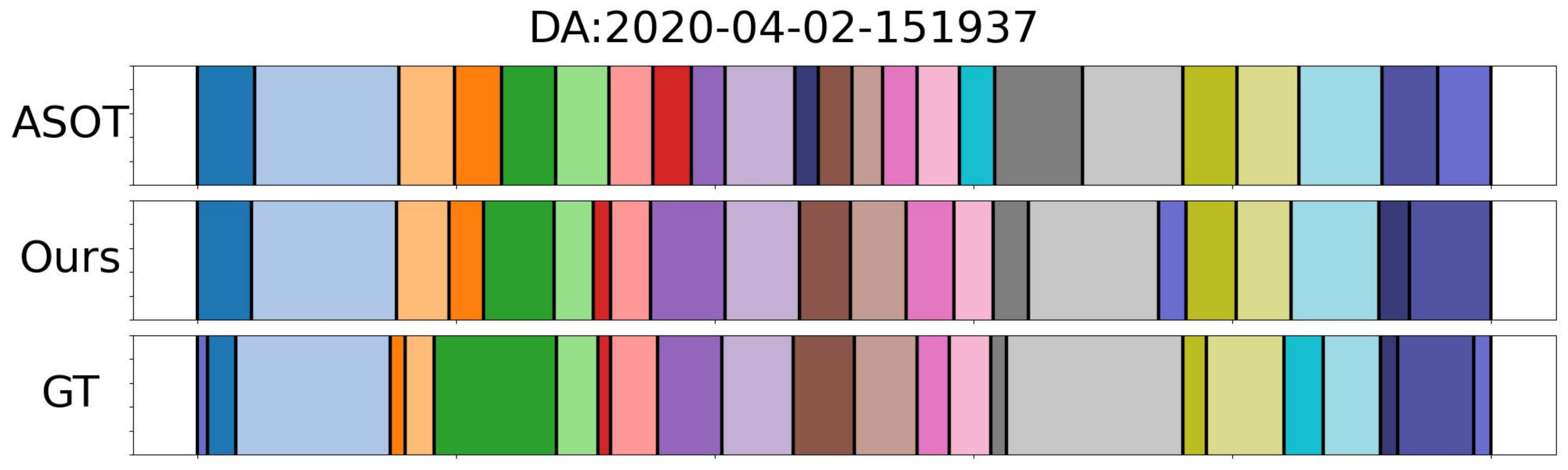}
    \caption{Desktop Assembly~\cite{kumar2022unsupervised}}
\end{subfigure}
 \caption{\textbf{Qualitative results.} Comparing ASOT~\cite{xu2024temporally}, PEOT (ours), and ground truth (GT) across different datasets and activities.}\label{fig:qualitative}
\end{figure*}

\begin{figure}[!t]
\centering
\begin{subfigure}[t]{0.48\linewidth}
    \centering
    \includegraphics[width=\linewidth]{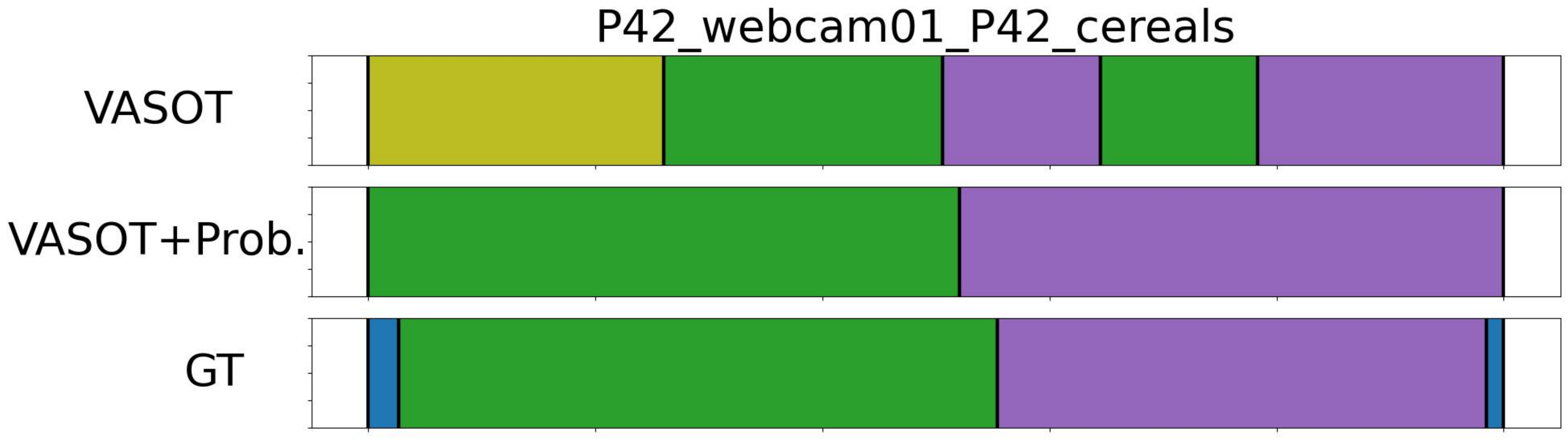}
\end{subfigure}
\hfill
\begin{subfigure}[t]{0.48\linewidth}
    \centering
    \includegraphics[width=\linewidth]
    {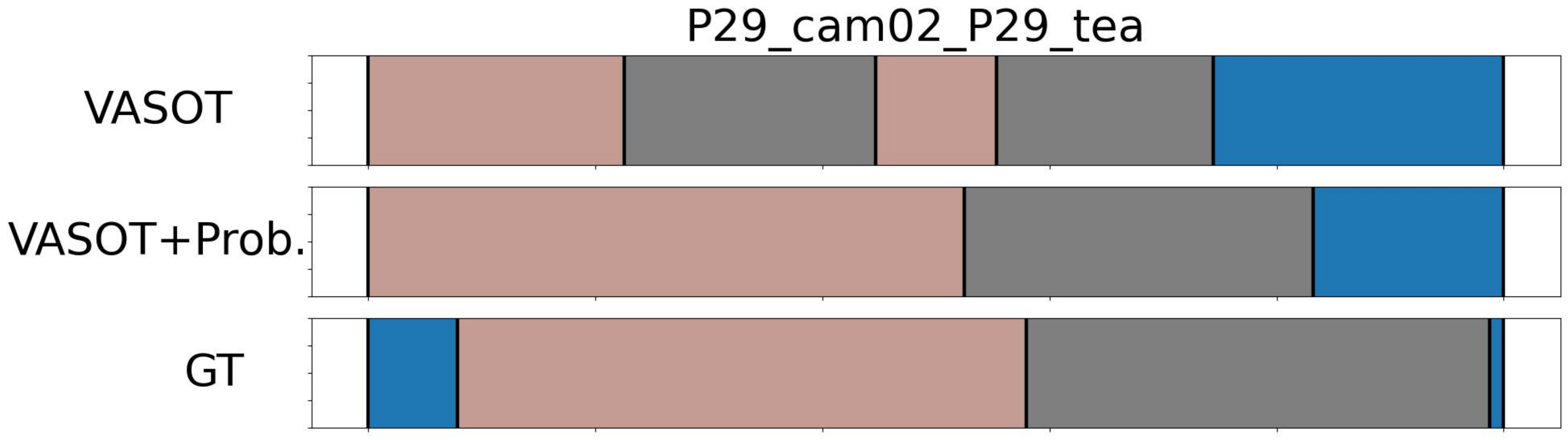}
\end{subfigure}
\caption{\textbf{Qualitative results.} Two examples when using VASOT~\cite{ali2025joint} as baseline. The examples are from the Breakfast~\cite{kuehne2014language} dataset.}
\label{fig:qualitative_vasot_prob}
\end{figure}

\subsection{Qualitative Results}\label{sec:qual}
We finally provide some qualitative segmentation results in~\cref{fig:qualitative}, comparing our approach to the ASOT baseline. In the example from the Breakfast dataset, our method is able to identify the long purple segment as opposed to the baseline approach which splits it into several segments. In the example from YTI, our method successfully segments very short actions. While ASOT finds good segments as well, more segments are not correctly clustered, \ie, not associated to the correct action. The black color indicates background frames. In the example from 50Salads, our method can detect reoccurring actions whereas ASOT does not recognize that some actions reoccur in this case. In the example from the Desktop Assembly dataset, the segments estimated by our approach are better aligned with the ground-truth. 

We further show some qualitative results for adding probabilistic embeddings to VASOT~\cite{ali2025joint} as baseline in~\cref{fig:qualitative_vasot_prob}. In these examples, VASOT generates additional segments for the green and gray cluster. With the probabilistic embeddings, the segments are better aligned with the ground truth. 

\vspace{-0.3cm}
\begin{figure}[!t]
\centering
\begin{subfigure}[t]{0.48\linewidth}
    \centering
    \includegraphics[width=\linewidth]{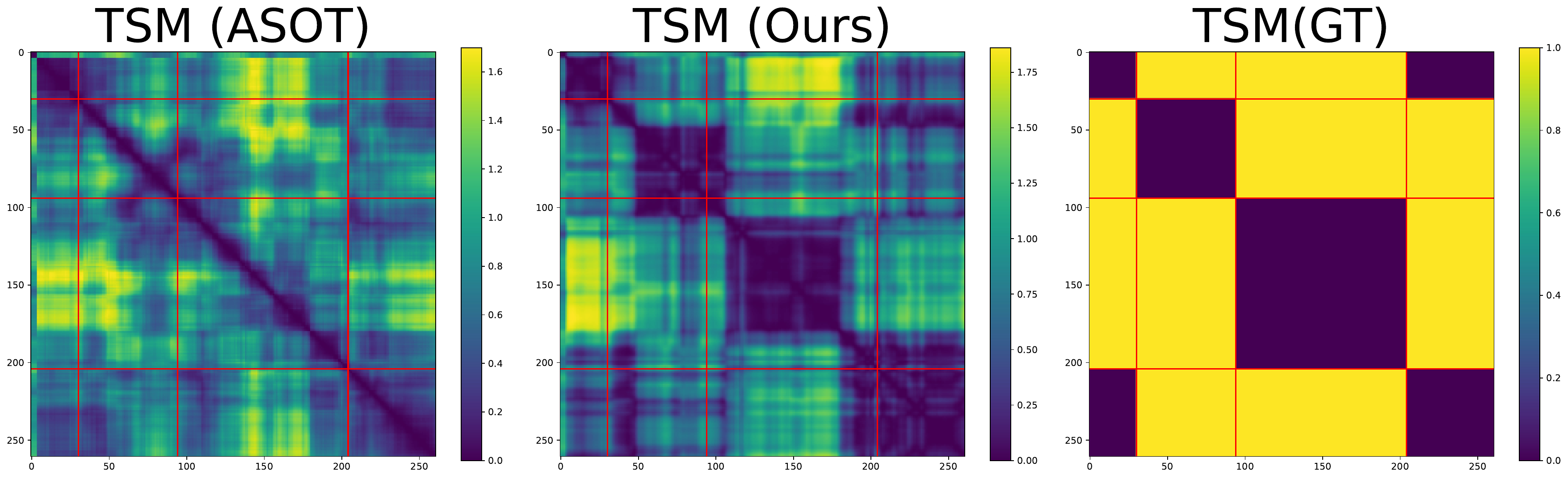}
    \caption{Breakfast: P12\_webcam01\_P12\_tea}
\end{subfigure}
\hfill
\begin{subfigure}[t]{0.48\linewidth}
    \centering
    \includegraphics[width=\linewidth]{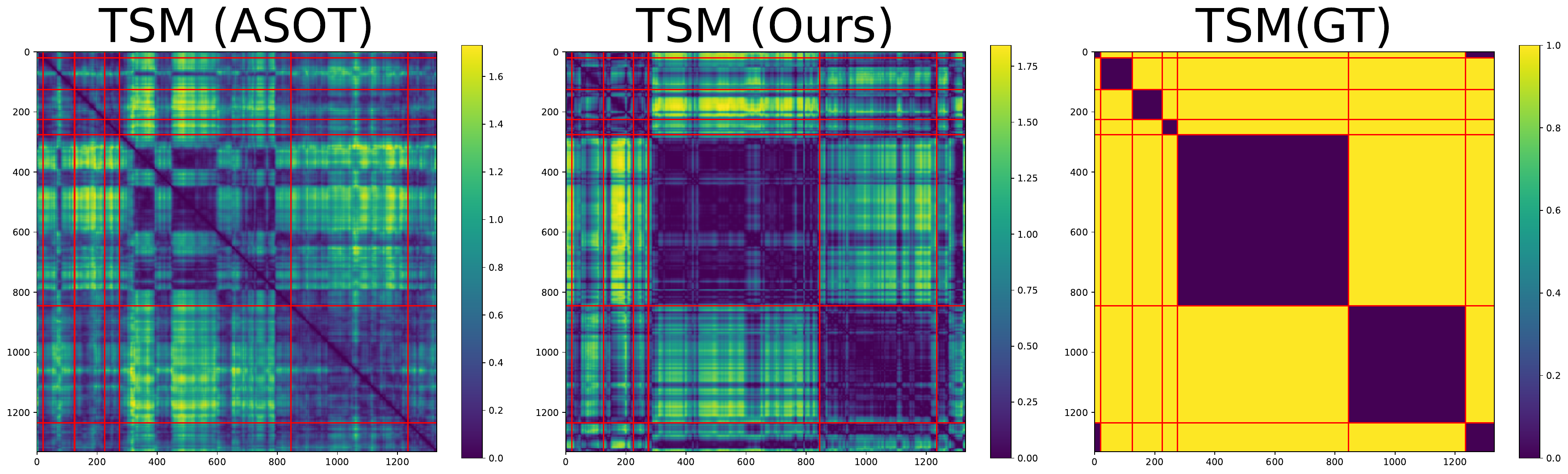}
    \caption{Breakfast: P15\_stereo01\_P15\_juice}
\end{subfigure}
\caption{\textbf{Qualitative results of the learned frame-wise representations.} Comparing ASOT~\cite{xu2024temporally} (left), our approach (middle), and ground truth (right) in terms of the temporal self-similarity matrix. Our approach learns better representations that are closer aligned to the ground truth segmentation.}
\label{fig:qualitative_tsm}
\end{figure}
\vspace{-0.3cm}

\subsubsection{Qualitative Results of Learned Frame Embeddings.}
We visualize the learned frame embeddings by calculating the temporal self-similarity matrix for a given video, where 1 minus the cosine similarity between every two frames is computed. We see in~\cref{fig:qualitative_tsm} that our approach gives better embeddings than ASOT~\cite{xu2024temporally}, which uses a deterministic embedding. In the first example our learned representations are more temporally coherent even for short segments, although it is slightly misaligned with the ground truth action boundaries. For the large blue squares in the ground truth matrix of the second example, ASOT shows stronger dissimilar patterns in the self-similarity matrix, which is an indicator that the learned representation focuses too much on subtle details that are not relevant for distinguishing actions.

\begin{figure*}[!t]
    \centering
    \begin{subfigure}[t]{0.48\linewidth}
        \centering
        \includegraphics[width=\linewidth]{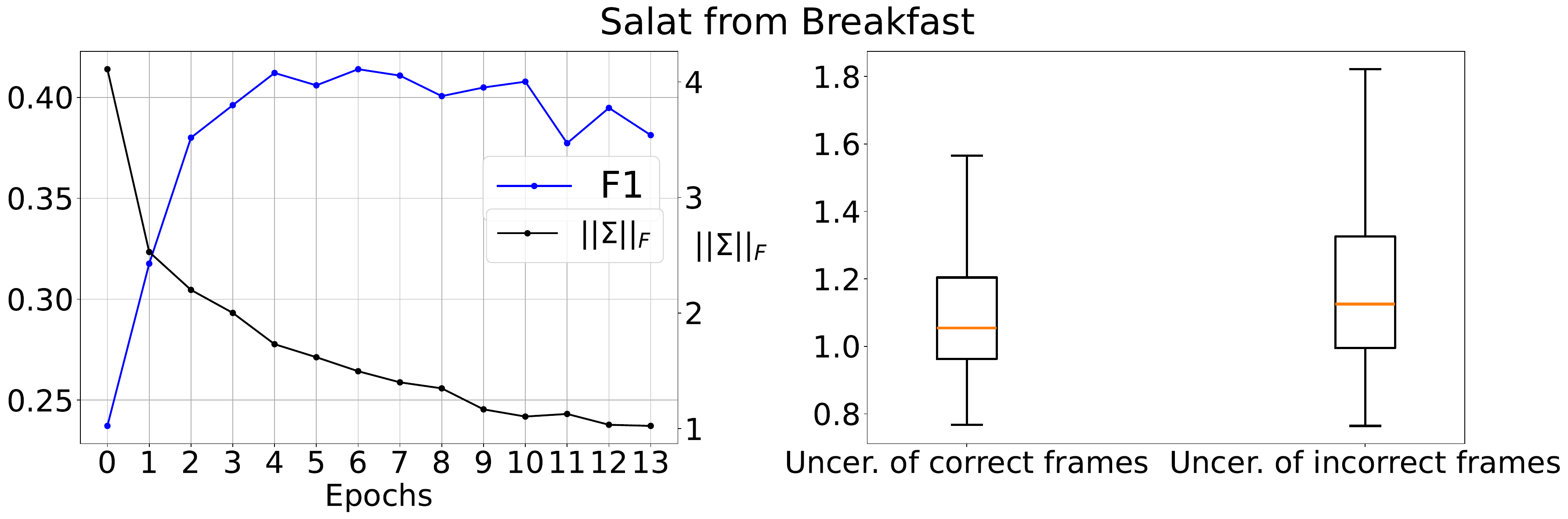}
        \caption{Uncertainty progress during training (left) and statistics of frame-wise uncertainty (right).}
        \label{fig:progress_uncertainty1}
    \end{subfigure}
    \hfill
    \begin{subfigure}[t]{0.48\linewidth}
        \centering
        \includegraphics[width=\linewidth]{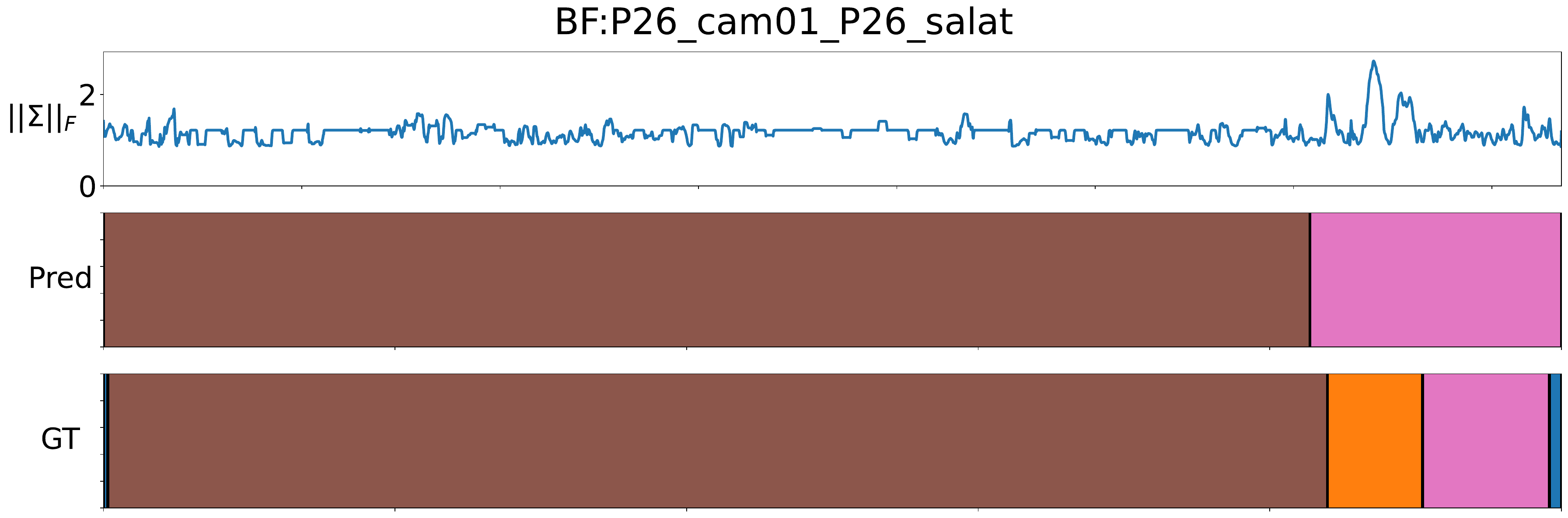}
        \caption{Qualitative result of segmentation and uncertainty estimation for one video of Breakfast.}
        \label{fig:qualitative_uncertainty}
    \end{subfigure}
    \caption{Analysis of the learned uncertainty.}
    \label{fig:uncertainty}
\end{figure*}

\begin{figure*}[!t]
    \centering   
    \includegraphics[width=0.95\linewidth]{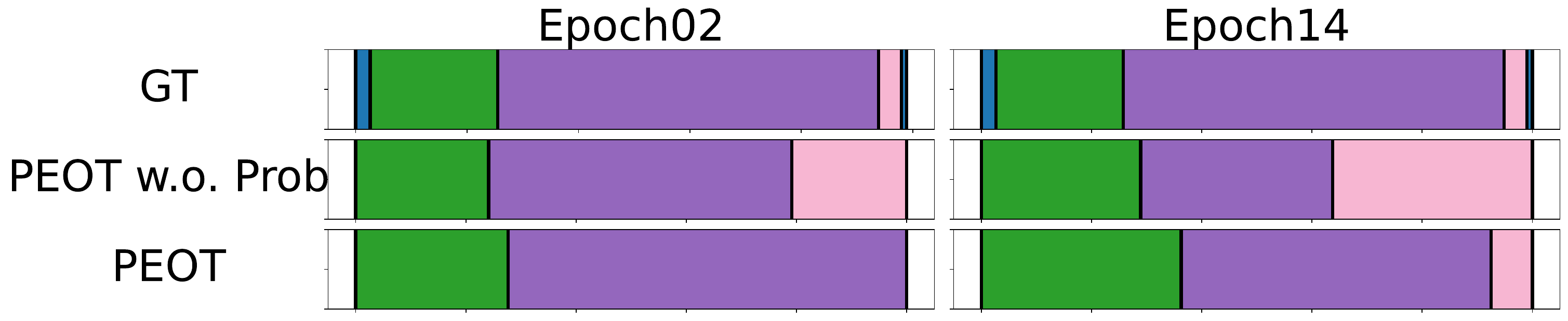}
    \caption{Difference between probabilistic embeddings and deterministic embeddings during training. 
    After 2 epochs, the approach using the deterministic embedding (row 2) recognizes the pink action, while the probabilistic embedding (row 3) misses it. After 14 epochs, the segmentation with the deterministic embedding only changes by shifting the boundaries between the segments whereas the pink action now also appears for the probabilistic embedding. The segmentation of the latter is now also closer to the ground truth.       
    The video is $\texttt{P52\_stereo01\_P52\_friedegg}$ taken from Breakfast.}
    \label{fig:qualitative_prob}
\end{figure*}

\subsection{Analysis of the Learned Uncertainty.}\label{sec:further_analysis}
We report the progress of the learned covariance during training in~\cref{fig:uncertainty}, using the Frobenius norm $||\Sigma||_F$ to measure the frame-wise uncertainty. In~\cref{fig:progress_uncertainty1}, we average the uncertainty across all frames of the $\texttt{Salat}$ activity of Breakfast. 
The uncertainty is high at the beginning of the training. As the training progresses, the learned features become better and the uncertainty decreases.  It might be beneficial to sample more often at the beginning of the training when the uncertainty is high, which we leave as a future work. We also observe that a low uncertainty is an indicator for overfitting.

We report a box plot (right of~\cref{fig:progress_uncertainty1}) for the uncertainties of all correctly predicted vs.\ all incorrectly predicted frames after training. It shows that the estimated uncertainty is in general higher for incorrectly predicted frames, but some incorrectly predicted frames have a low uncertainty. This is visualized in~\cref{fig:qualitative_uncertainty}. For the missed orange segment, the peaks of the uncertainty are higher compared to the correctly detected segments, but the uncertainty is not high for all frames within the orange segment.

We further plot the evolution of the results for two different training epochs in~\cref{fig:qualitative_prob}. The example shows that the deterministic embedding provides a better segmentation at the early epoch, but the probabilistic embedding changes more during training. At the later epoch, the segmentation of the probabilistic embedding is closer to the ground-truth. 

\section{Conclusion}
In this work, we proposed learning probabilistic instead of deterministic embeddings for frame representations, which brings a novel perspective to unsupervised temporal action segmentation. We incorporated the approach into two state-of-the-art baselines, namely ASOT~\cite{xu2024temporally} and VASOT~\cite{ali2025joint}, and evaluated it on four challenging benchmarks. The results showed that the probabilistic embeddings consistently improved MoF and F1-score over all datasets. Compared to the state of the art, our approach achieves the best performance for 9 out of 12 metrics/datasets. Future work can make the pseudo label generation process differentiable~\cite{li2022learning} and the probabilistic embeddings might be also useful for other tasks like action anticipation~\cite{zatsarynna2024gated}.        
\label{sec:conclusion}

\section*{Acknowledgements}
The work has been supported by the project iBehave (receiving funding from the programme ``Netzwerke 2021'', an initiative of the Ministry of Culture and Science of the State of Northrhine Westphalia) and the ERC Consolidator Grant FORHUE (101044724). 
The authors would like to thank Elena Bueno-Benito, Federico Spurio and Yazan Abu Farha for helpful discussions.

%
%
\bibliographystyle{splncs04}
\bibliography{main}

\clearpage
\appendix
\title{Learning Probabilistic Embeddings for Unsupervised Action Segmentation Supplementary Materials}

\titlerunning{Learning Probabilistic Embeddings for Unsupervised Action Segmentation}

\author{Shuai Li\inst{1} \and
Duc Manh Vu\inst{1} \and
Juergen Gall\inst{1,2}}

\authorrunning{S.~Li et al.}

\institute{University of Bonn, Germany \and
Lamarr Institute for Machine Learning and Artificial Intelligence, Germany
\email{\{lishuai,gall\}@iai.uni-bonn.de}}

\maketitle

\section{Analysis of Probabilistic Embeddings}
\begin{figure}[!h]
    \centering
    \includegraphics[width=0.7\linewidth]{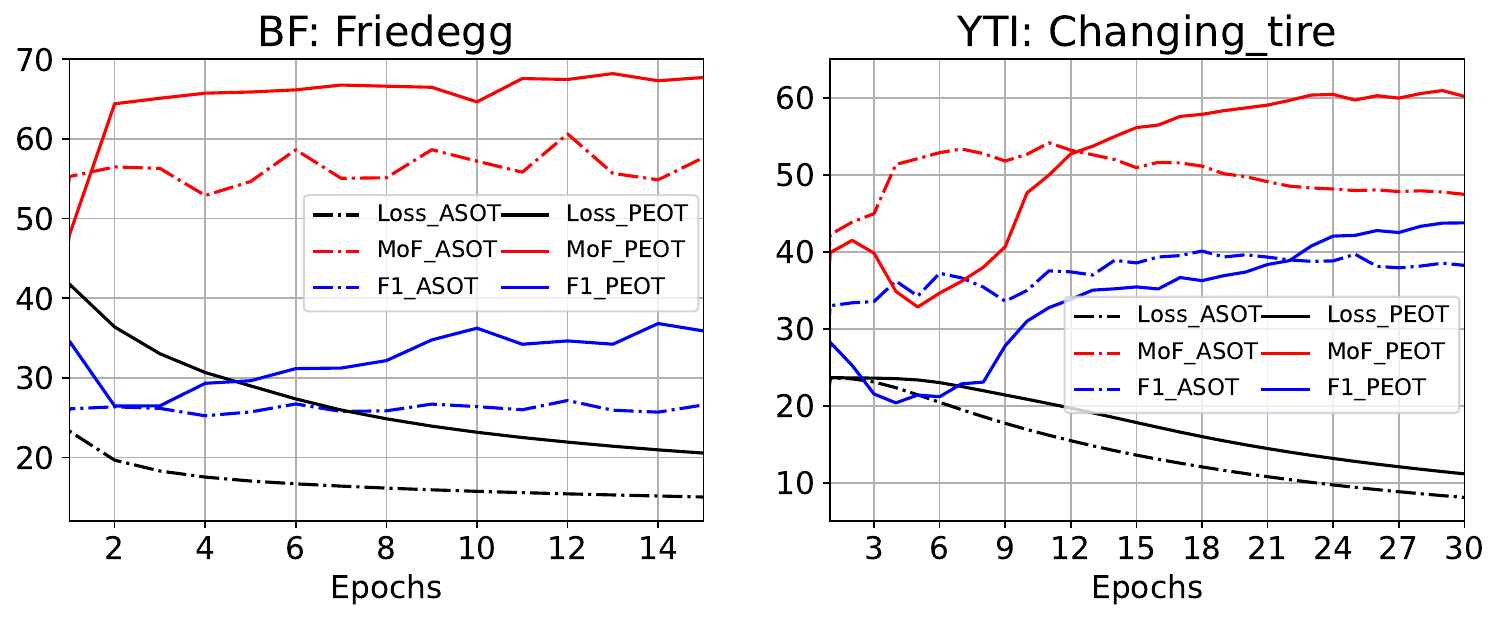}
    \caption{Comparing for ASOT~\cite{xu2024temporally} (dashed) and our approach PEOT (solid) the loss curves (black) and evaluation metrics MoF (red) and F1 (blue) for videos of the friedegg and changing\_tire activity from the Breakfast~\cite{kuehne2014language} and Youtube Instr.~\cite{alayrac2016unsupervised} dataset, respectively.  We scale the loss for better visualization. }
    \label{fig:loss_and_metrics}
\end{figure}

While the quantitative results in the paper already show the benefit of learning probabilistic embeddings for unsupervised action segmentation, we further analyze its impact on the training. \cref{fig:loss_and_metrics} shows the loss (black curve) for the ASOT baseline with deterministic embeddings and our approach with probabilistic embeddings. The left figure plots loss, MoF, and F1-score for the $\mathtt{Friedegg}$ activity from Breakfast~\cite{kuehne2014language}. ASOT achieves a lower loss but both the MoF and F1-score saturate as the training progresses.  
In contrast, despite the higher loss, the MoF and F1-score of our approach keep increasing throughout the training. Similarly, the right figure shows the plot for the $\mathtt{Changing\_tire}$ activity from Youtube Instr.~\cite{alayrac2016unsupervised}. ASOT achieves better MoF and F1 at the beginning of the training but it does not improve much. MoF even decreases after 15 epochs. On the other hand, our approach improves MoF and F1-score as training continues. 

\vspace{-0.8cm}
\section{Sensitivity Analysis}
\begin{figure}[!h]
\centering
    \includegraphics[width=\linewidth]{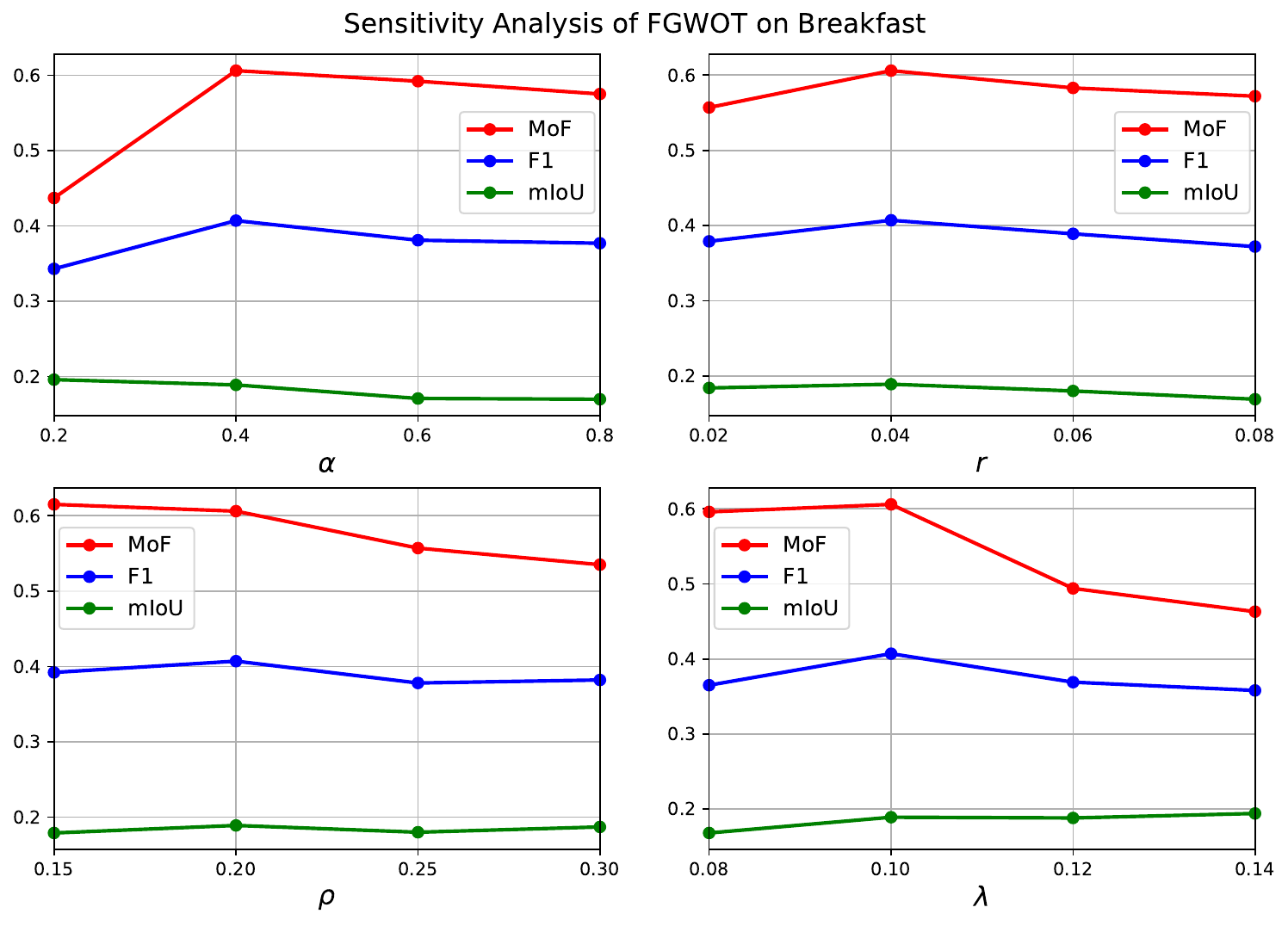}
\caption{Sensitivity analysis of OT hyper-parameters.}
\label{fig:sensitivity}
\end{figure}

For the parameters of the fused Gromov-Wasserstein optimal transport, we follow ASOT, using $\alpha_\text{train}=0.4$, $r=0.04$, $\rho=0.2$, and $\lambda_\text{train}=0.1$ for Breakfast, $\alpha_\text{train}=0.3$, $r=0.02$, $\rho=0.2$, and $\lambda_\text{train}=0.12$ for Youtube Instr., $\alpha_\text{train}=0.3$, $r=0.02$, $\rho=0.05$, and $\lambda_\text{train}=0.1$ for 50Salads (Eval), and $\alpha_\text{train}=0.3$, $r=0.02$, $\rho=0.25$, and $\lambda_\text{train}=0.16$ for Desktop Assembly. In \cref{fig:sensitivity}, we report the sensitivity of these parameters on Breakfast. MoF and F1 benefit from an increased $\alpha_\text{train}$ as more temporal consistency is enforced, but this leads to a slightly decrease in mIoU. We thus use $\alpha_\text{train}=0.4$ during the experiments. The highest performance is achieved when $r=0.04$ as larger $r$ leads to over-smoothing. Increasing $\rho$ would result in a decrease of MoF and F1 since the performance on activities involving repetitive actions would drop, therefore $\rho=0.2$ is used. We also observe a notable performance drop with increased unbalanced weight $\lambda_\text{train}$ although mIoU becomes better. For a tradeoff, we set $\lambda_\text{train}=0.1$. 


\section{Training Time}
We use the same OT formulation as ASOT~\cite{xu2024temporally} and conducted all experiments on a single NVIDIA GeForce RTX 3090 GPU (24GB) with CUDA 11.8. Compared to ASOT, our method PEOT only contains one additional GCN layer followed by a sampling procedure during training. The results in~\Cref{tab:flops_comparison} show that our method increases the training time per epoch by 2 seconds, which is a moderate increase. 
Compared to our approach PEOT, VASOT~\cite{ali2025joint} consumes twice of the time per training epoch due to the use of video-to-video alignment. 

\begin{table}[!ht]
\begin{center}
\caption{Average per-epoch training time on Breakfast~\cite{kuehne2014language}.}
\label{tab:flops_comparison}
\begin{tabular}{l c}
 \hline
 Methods & Time (s) \\ 
 \hline
ASOT~\cite{xu2024temporally} & \textbf{4.494} \\ 
VASOT~\cite{ali2025joint} & 13.410 \\ 
PEOT (Ours) &  6.578 \\
\hline
\end{tabular}
\end{center}
\end{table}

\section{Segmentation Performance over Multiple Runs}
We study the segmentation performance over multiple runs. To do so, we run our method as well as ASOT~\cite{xu2024temporally} and VASOT~\cite{ali2025joint} for 3 different random seeds. On Breakfast, the standard deviation for F1 is 0.7, 1.4, and 2.1 for Ours, ASOT~\cite{xu2024temporally}, and VASOT~\cite{ali2025joint}, respectively. For mIoU, it is 0.4, 0.6, and 0.9, respectively.



%
\end{document}